%% file: main.tex
\pdfoutput=1
\documentclass[11pt]{article}

\usepackage{acl}

\usepackage{times}
\usepackage{latexsym}

\usepackage[T1]{fontenc}
\usepackage[utf8]{inputenc}

\usepackage{microtype}

\usepackage{inconsolata}

\usepackage{amsmath}
\usepackage{amssymb}
\usepackage{mathtools}
\usepackage{amsthm}
\usepackage{bbm}
\usepackage{booktabs}
\usepackage{subfigure}
\usepackage{multirow}
\usepackage{wrapfig,lipsum}
\usepackage{paracol}
\usepackage{adjustbox}
\usepackage{multicol}
\usepackage{enumitem}
\usepackage{subcaption}
\usepackage{colortbl}
\usepackage{comment}

\newcommand\blfootnote[1]{%
  \begingroup
  \renewcommand\thefootnote{}\footnote{#1}%
  \addtocounter{footnote}{-1}%
  \endgroup
}

\title{Semantic Token Reweighting for Interpretable and Controllable \\ Text Embeddings in CLIP}

\author{Eunji Kim$^{1,2\ast}$ \hspace{1em} Kyuhong Shim$^2$ \hspace{1em} Simyung Chang$^{2{\dag}}$ \hspace{1em} Sungroh Yoon$^{1,3{\dag}}$ \\
{$^1$Department of Electrical and Computer Engineering, Seoul National University} \\
{$^2$Qualcomm AI Research$^{\ddagger}$, Qualcomm Korea YH} \\
{$^3$Interdisciplinary Program in Artificial Intelligence, Seoul National University}\\
{\texttt{\{kce407, sryoon\}@snu.ac.kr}} \\
{\texttt{\{kshim, simychan\}@qti.qualcomm.com}}}

\begin{document}
\maketitle

\blfootnote{\hspace{-1.8em}$^\ast$ Work done during an internship at Qualcomm Technologies, Inc. \\$^\ddagger$Qualcomm AI Research, an initiative of Qualcomm Technologies, Inc.\\
$^\dag$Corresponding authors} 

\input{texts/00.abstract}
\input{texts/01.introduction}
\input{texts/03.method}
\input{texts/04.experiments}
\input{texts/02.related_work}
\input{texts/05.conclusion}

\input{texts/06.limitation}

% \section*{Acknowledgments}

% \bibliographystyle{acl_natbib}
\bibliography{custom}

\newpage
\input{texts/appendix}

\end{document}

%% file: texts/00.abstract.tex
\begin{abstract}
A text encoder within Vision-Language Models (VLMs) like CLIP plays a crucial role in translating textual input into an embedding space shared with images, thereby facilitating the interpretative analysis of vision tasks through natural language. Despite the varying significance of different textual elements within a sentence depending on the context, efforts to account for variation of importance in constructing text embeddings have been lacking. We propose a framework of Semantic Token Reweighting to build Interpretable text embeddings (SToRI), which incorporates controllability as well. SToRI refines the text encoding process in CLIP by differentially weighting semantic elements based on contextual importance, enabling finer control over emphasis responsive to data-driven insights and user preferences. The efficacy of SToRI is demonstrated through comprehensive experiments on few-shot image classification and image retrieval tailored to user preferences.
\end{abstract}

%% file: texts/01.introduction.tex
\section{Introduction}\label{sec:intro}

\input{tabs/fig_diagram}

As artificial intelligence (AI) systems based on deep learning models grow in application in our daily lives, their black box nature raises issues of transparency, resulting in a demand for enhanced interpretability to promote trust in AI systems~\citep{murdoch2019definitions,li2022interpretable}. Consequently, research efforts have been focused on making the systems' decision-making processes more human-understandable through various explanatory methods~\citep{simonyan2013deep,kim2018interpretability,goyal2019counterfactual,wu-mooney-2019-faithful}.
Among the various forms of explanation, natural language has emerged as an excellent medium due to its human-friendly nature and adeptness in managing high-level abstractions~\citep{kayser2021vil,sammani2022nlx}.
These advantages have led to a growing interest in leveraging natural language for interpreting vision tasks~\citep{hendricks2021generating,yang2023language}.

To facilitate the use of natural language in vision tasks, Vision-Language Models (VLMs) like CLIP~\citep{radford2021learning} are commonly deployed to bridge visual information and its linguistic interpretation~\cite{yuksekgonul2023posthoc,yang2023language,oikarinen2023labelfree}.
CLIP consists of two encoders that translate images and texts into embeddings that coexist in a shared space, enabling vision tasks to be conducted and understood through natural language.

Natural language sentences often carry multiple implications, with varying levels of significance that can change based on the desired outcome, even if the text remains unchanged.
Selectively emphasizing certain information relevant to a task can aid in conducting and understanding the task.
For instance, when differentiating given images of a `great grey owl' from other groups using the description `a large owl with big yellow eyes', emphasis on `eyes' may better represent the group of images, indicating that `eyes' is significant (see examples of image classification in Figure~\ref{fig:diagram}).
Similarly, when searching for images using the query `a castle surrounded by trees,' the preference on `trees' relative to `a castle' could differ based on user intent, and retrieval reflecting this can yield the desired search results ({see examples of retrieved images in Figure~\ref{fig:diagram}}).
While there have been attempts to modulate focus in image and text generation~\citep{ge2023expressive,zhang2023prompt,zhang2024tell}, fine-tuning the importance of specific text elements in CLIP's text embeddings remains relatively unexplored.
This paper endeavors to create text embeddings that can incorporate a varying controlled importance of each semantic element within a sentence, thereby enhancing the representativeness of text embedding for images in interpretable way.

To meet our objective, we introduce \textbf{SToRI} (\underline{S}emantic \underline{To}ken \underline{R}eweighting for \underline{I}nterpretable text embeddings). SToRI adjusts the importance of each semantic element during text embedding extraction in CLIP by assigning a weight to each element, denoting its significance, which modulates the self-attention mechanism in text encoding. This allows the final text embedding vector to reflect the desired emphasis on specific elements, enhancing representativeness for vision tasks without requiring new modules. Since the emphasis remains within an interpretable space, SToRI also enables the interpretative analysis of vision tasks using natural language.

Our SToRI framework offers two ways of tailoring text embeddings: data-driven and user-driven. The data-driven approach derives token weights from training on dataset, optimizing text embeddings for image classification and revealing interpretable insights (see the orange path in Figure~\ref{fig:diagram}). The user-driven approach allows users to set weights for each semantic token, customizing the text embedding to fit their preferences (see the green path in Figure~\ref{fig:diagram}). We demonstrate these enhancements through two vision tasks with CLIP: few-shot classification and image retrieval.

To summarize, our main contributions are:
\begin{itemize}
\item We propose a novel framework to differentiate the importance of textual information during the construction of text embeddings with CLIP for vision tasks.
\item Our method can build improved text classifiers in few-shot learning tasks while offering new interpretability insights.
\item We demonstrate the controllability of our method, specifically customization of semantic emphasis, and its utility in image retrieval tasks using a new metric.
\end{itemize}

%% file: tabs/fig_diagram.tex
\begin{figure*}[t]
\begin{center}
\centerline{\includegraphics[width=\linewidth]{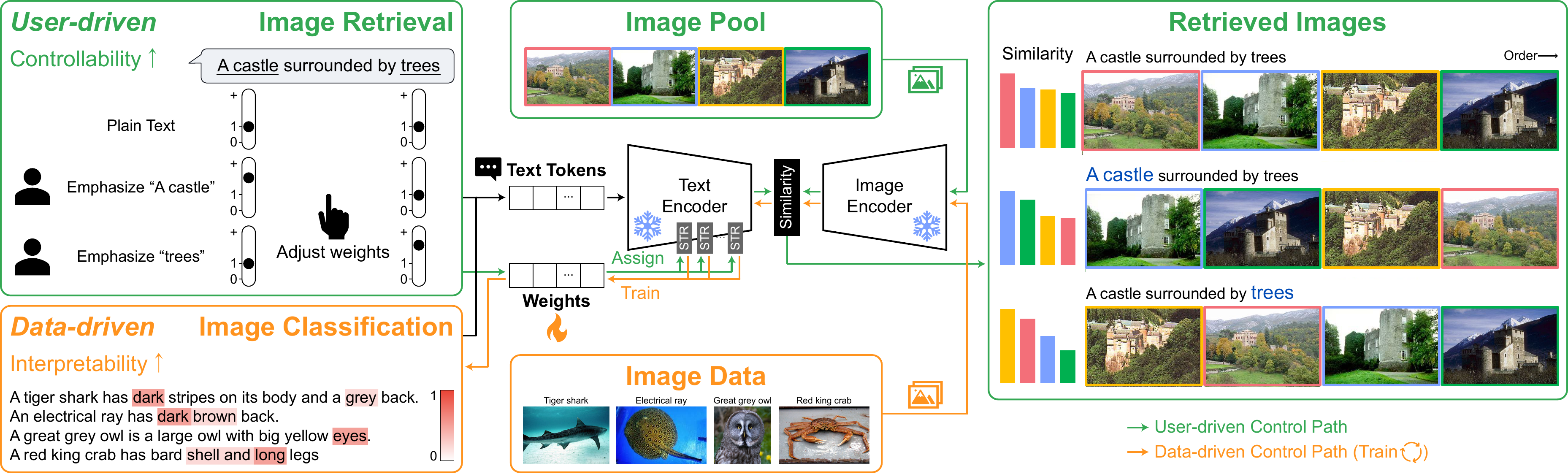}}
\caption{System diagram of SToRI. SToRI facilitates \textit{data-driven control} through interpretable weight optimization in the semantic space, enhancing the classification performance of image data. It also enables \textit{user-driven control} over multiple images by allowing fine-grained manipulation of the text prompts. Weights affect text embeddings via semantic token reweighting (STR).}
\label{fig:diagram}
\end{center}
\end{figure*}

%% file: texts/03.method.tex
\section{Preliminary: Text embeddings in CLIP}\label{sec:preliminary}
The text encoder of CLIP~\citep{radford2021learning}, which utilizes a transformer-based architecture, transforms a given text prompt into a single vector through the following process.
Initially, a given text prompt is converted into a sequence of text tokens $\{x_i\}_{i=1}^{N}$, where $N$ represents the number of the text tokens. Tokens indicating the start and end, \texttt{[SOS]} and \texttt{[EOS]} tokens, are appended at the beginning and the end of the sequence of tokens, resulting in the extended series $\{x_i\}_{i=0}^{N+1}$, with $x_0$ and $x_{N+1}$ representing the \texttt{[SOS]} and \texttt{[EOS]}, respectively. Each text token is then converted into an embedded input token, and positional embedding is added, resulting in the input embedding for the first transformer block $\{z_i^0\}_{i=0}^{N+1}$.
For the $l$-th block of the encoder, the input tokens can be represented as $Z^{l-1}=[z_0^{l-1}, ..., z_{N+1}^{l-1}]$. The output tokens from the $l$-th block is given by:
\begin{equation}
    Z^{l}=\text{Block}^l(Z^{l-1}),
\end{equation}
where $l \in [1,L]$ with the encoder consisting of $L$ blocks.
Each block contains a multi-head self-attention mechanism. First, $Z^{l-1}$ is projected into the query $Q$, key $K$, and value $V$. Then, the attention process is performed as follows:
\begin{equation}\label{eq:attention}
\begin{aligned}
    &\text{Attention}(Q, K, V)=AV,\\
    &\text{s.t.}~A=\text{softmax}(QK^T).
\end{aligned}
\end{equation}
Scaling and masking operations are omitted for simplicity. Through the attention mechanism, tokens influence each other, and the values of $A$ represent the extent to which they influence one another~\cite{vaswani2017attention}.
In general, the final output text embedding of the \texttt{[EOS]} token encapsulates the full semantic meaning of the text prompt. This embedding is compared with image embeddings to assess the degree of correspondence with images once it has been projected into a multi-modal embedding space.

A pre-trained CLIP model is commonly employed for image classification, where given an image, it computes similarity scores with class names, which become logits. To adapt the model to a specific dataset, fine-tuning is performed by minimizing the cross-entropy loss as follows:
\begin{equation}\label{eq:loss_ce_preliminary}
    \mathcal{L}=L_\text{CE}(y, \text{sim}(\phi_T, \phi_I)/\tau),
\end{equation}
where $\phi_T$ and $\phi_I$ represent output text and image embeddings from two encoders, respectively, $\tau$ is a temperature factor, and $y$ is a target class.

\section{Method}\label{sec:method}
We propose SToRI, a novel framework that encodes a given text prompt into a single text embedding vector using CLIP by varying the importance of different textual elements through data-driven and user-driven controls. In Section~\ref{sec:method_str}, we elaborate on semantic token reweighting, which involves modifying the attention given to individual tokens within the text encoding process based on their respective weights. In Section~\ref{sec:method_control}, we present two methods for determining these weights.

\subsection{Semantic Token Reweighting}\label{sec:method_str}
In natural language processing, a given text is tokenized prior to encoding, resulting in one or more tokens. Consequently, to emphasize or de-emphasize a particular semantic element, one must focus on the corresponding tokens. Henceforth, our discussion will center on the process of reweighting in terms of these tokens.

Given a sequence of text tokens $\{x_i\}_{i=1}^N$, we first define a sequence of weights $\{w_i\}_{i=1}^{N}$, where $w_i$ is the level of significance of token $x_i$. Note that $w_i=1$ indicates a typical weight in common situations, where $x_i$ is neither emphasized nor de-emphasized.
Our goal is to modulate the impact each token has on the final output embedding of the text prompt.
As elaborated in Section~\ref{sec:preliminary}, tokens interact with each other through attention mechanisms. Each token generates its embedding by referencing other tokens, including itself, in proportion to the attention scores. Consequently, as the attention score of a specific token increases, its influence on the text embedding becomes more substantial. Therefore, we directly multiply the weights $\{w_i\}_{i=1}^{N}$ to amplify original attention values proportionally. From Eq.~\eqref{eq:attention}, the weighted attention scores can be reformulated as follows:
\begin{equation}
    \hat{a}_{m, n}=\frac{w_n\exp{(q_m k_n^T})}{\sum_j{w_j\exp{(q_m k_j^T}})},
\end{equation}
where $\hat{a}_{m,n}$ represents attention value for $n$-th value token to be attended by $m$-th query token. $q_m$ and $k_n$ represent vector elements of $Q$ and $K$, respectively.
Through this process, we can selectively enhance the influence of particular tokens during the attention process by simply changing the corresponding weights.
 
The reweighting process is applied to all blocks following a certain block. Experimentally, we confirm that the effects are similar regardless of starting from any intermediate block. Please refer to Appendix~\ref{sec:appendix_exp_position} for further details.

\input{tabs/fig_interpretation1}

\subsection{Strategies to Control}\label{sec:method_control}
There are two approaches to determine weights for tokens: user-driven and data-driven controls.

\textbf{Data-driven control} determines weights by learning from data. This approach is suitable when data is available and we want to obtain text embeddings that align closely with the data. As shown with the orange path in Figure~\ref{fig:diagram}, an illustrative task where this can be effectively applied is image classification. In image classification, weights are trained using Eq.~\eqref{eq:loss_ce_preliminary}, where $\phi_T$ is obtained with $\hat{a}_{i, j}$, allowing only $\{w_i\}_{i=1}^{N}$ to be updated. Since the weights are trained to build text embeddings that correspond well to image belonging to their corresponding classes, we can interpret which textual information prominently stands out in the image data with the weights.

\textbf{User-driven control} applies to scenarios where the user assigns weights to each token. This method allows user to determine a particular textual information to be emphasized or de-emphasized according to their intentions, thereby influencing the resulting text embeddings.
{The green path in Figure~\ref{fig:diagram} presents examples of preference-based image retrieval, an application in the user-driven control. Users may initially set a text prompt and then progressively amplify the weight of keywords perceived as more crucial, assess the resulting arrangement, and refine their selection accordingly.}

\input{tabs/fig_interpretation2}

%% file: tabs/fig_interpretation1.tex
\begin{figure*}[t]
\begin{center}
\centerline{\includegraphics[width=0.87\linewidth]{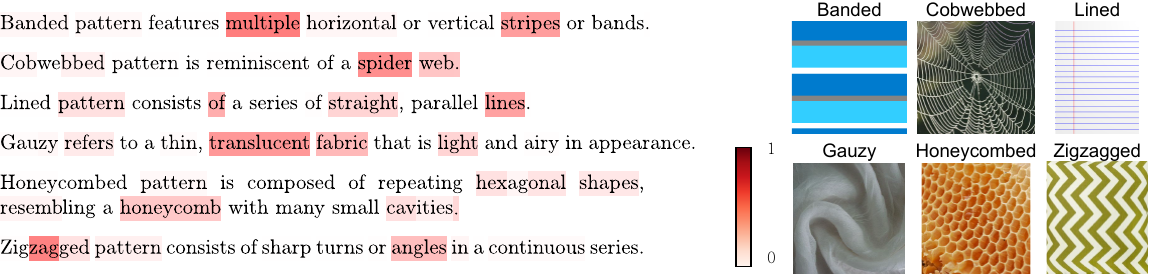}}
\caption{Text prompts and corresponding weights are provided as examples after training. The intensity of the red shading reflects the weight assigned, with darker shades indicating higher weights. For visualization, the weights are normalized to sum up 1. The figures on the right display an example image for each class.}
\label{fig:result_interpretation_dtd}
\end{center}
\end{figure*}

%% file: tabs/fig_interpretation2.tex
\begin{figure*}[htb!]
\begin{center}
\centerline{\includegraphics[width=0.9\linewidth]{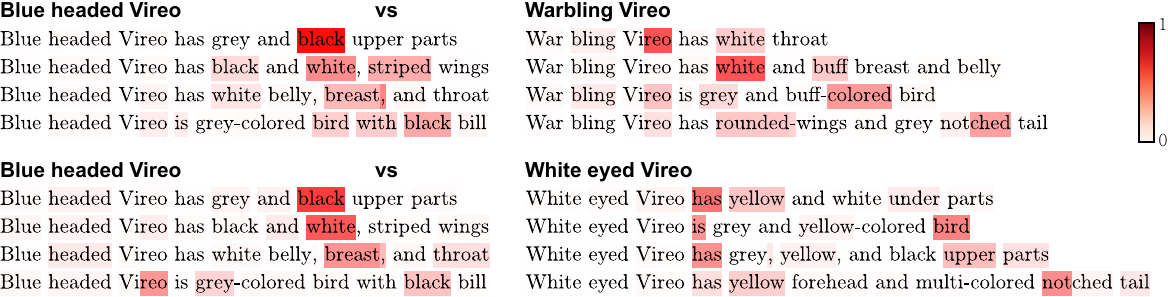}}
\caption{Text prompts and their corresponding weights are presented after training with the CUB dataset. The more intense the shade of red, the greater the weight assigned. In each scenario, the text classifier is trained to discriminate two classes. The weights for the same text prompts vary depending on the class to be distinguished.}
\label{fig:result_interpretation_cub}
\end{center}
\end{figure*}

%% file: texts/04.experiments.tex
\section{Experiments}\label{sec:exp}
We evaluate SToRI on two representative vision tasks, few-shot image classification and preference-based image retrieval. In few-shot image classification, weights are determined via data-driven control and provide interpretation on the trained classifier. Evaluation on preference-based image retrieval demonstrates controllability of SToRI via user-driven control.

\subsection{Classification with Data-driven Control}
We train weights that best represent each dataset for the image classification task. We first show interpretation with trained weights and then evaluate few-shot classification performance of trained text classifier.

\subsubsection{Experimental Setup}

\paragraph{Datasets}
We use DTD~\cite{cimpoi2014describing} and CUB~\cite{wah2011caltech} datasets for analysis on interpretation.
We use various benchmarks for few-shot learning \textit{i.e.}, ImageNet~\cite{deng2009imagenet}, DTD~\cite{cimpoi2014describing}, SUN397~\cite{xiao2010sun}, Flowers102~\cite{nilsback2008automated}, Caltech101~\cite{fei2004learning}, and Food101~\cite{bossard2014food}.

\paragraph{Text Prompts}
We use text descriptions for each class which are provided by CuPL~\citep{pratt2023does}.
For the ImageNet and SUN397 datasets, due to the large number of total prompts, we use 10 text prompts for each class, selected based on their similarity with training set. We average the text embeddings from multiple text prompts to build one text embedding for each class. We refer the text embedding for image classifier as a text classifier.

\paragraph{Model}
The experiments are conducted using CLIP and MetaCLIP ViT-L/14, with reweighting applied from the 7th block onward.

\paragraph{Implementation Details}
We set the logarithm of the weight as the parameter to be trained in order to constrain the weights to non-negative values. Each text prompt has its own individual set of weights.

\paragraph{Training Details}
Following TaskRes~\cite{yu2023task}, we evaluate our method by training with 1/2/4/8/16 examples (shots) per class from the training sets, respectively.
We follow the data split outlined in CoOp~\cite{zhou2022learning}, conducting tests on the official test set of each dataset and the validation set of the ImageNet dataset.
1/2/4-shot training is done with 100 epoch and the other is done with 200 epoch for all datasets.

For further details, please refer to Appendix~\ref{sec:appendix_exp_detail_data}.

\subsubsection{Interpretability}

\paragraph{Interpretation with Trained Weights}
After training for an image classification task, we analyze the trained weights. Figure~\ref{fig:result_interpretation_dtd} presents examples of text prompts and the corresponding trained weights for each token within the DTD dataset.
We have crafted the text prompts. We can discern that \textit{banded} is associated with an emphasis on words like \texttt{multiple} and \texttt{stripes}.
For \textit{gauzy}, terms such as \texttt{translucent} and \texttt{light} are emphasized, and \textit{cobwebbed} are notably associated with the word \texttt{spider web}. As illustrated by the images corresponding to each category, high weight values are assigned to important semantic tokens. This shows that SToRI can learn text embeddings that effectively represent the data in a data-driven control context, and the trained weights can offer novel insights for interpretation.

\paragraph{Does Optimization Occur in Interpretable Space?}
To ensure interpretability of text embeddings through data-driven control optimization, we conduct two experiments: an analysis on trained classifiers with different class compositions and an assessment of the effect of nonsensical text tokens.

The role of classifier is to distinguish one class from others. Thus, even for classifiers within the same class, the critical distinguishing features can vary depending on the alternative categories being compared. Figure~\ref{fig:result_interpretation_cub} shows two text classifiers trained on the CUB dataset for two distinct pairs: \textit{Blue headed Vireo} versus \textit{Warbling Vireo}, and \textit{Blue headed Vireo} versus \textit{White eyed Vireo}.
The text prompts for each class are generated with the attribute labels from the dataset. When contrasting \textit{Blue headed Vireo} with the \textit{Warbling Vireo}, \texttt{striped} is attributed a high weight. However, when distinguished from the \textit{White eyed Vireo}, the weight on \texttt{striped} becomes low and \texttt{grey} is attributed a high weight. Note that \textit{White eyed Vireo} also has striped wings.
These terms highlight the key distinctions between each pair of classes.

\input{tabs/tab_fewshot_textdescription}

To evaluate whether simply increasing the number of trainable parameters—specifically through the addition of nonsensical tokens—would enhance performance, we compare 16-shot classification results with and without the inclusion of such tokens. If performance improvements stemmed solely from the increased number of trainable parameters, rather than from optimizing attention on semantically meaningful tokens, we would expect a boost in performance when nonsensical tokens are added.
We randomly sample five tokens from the set of three rare tokens~\citep{ruiz2023dreambooth}, namely \texttt{`sks'}, \texttt{`pll'}, and \texttt{`ucd'}, and add them to the end of all the original texts from CuPL. The inclusion of rare tokens does not contribute meaningful information to build a text classifier; it simply extends the number of tokens and trainable parameters. As shown in Table~\ref{tab:compare_text_few_perf}, the performance with rare tokens does not exceed the baseline without their inclusion. This demonstrates that adoption of the tokens without semantic meaning does not contribute to performance improvement.
These findings support that data-driven control, achieved through attention modulation for tokens with semantic meaning, facilitates the creation of text embeddings that effectively represent the data, thereby ensuring the interpretability of text embeddings.

\input{tabs/tab_fewshot}
\subsubsection{Few-shot Classification Performance}
To evaluate the capability of the text classifier obtained through SToRI to perform few-shot image classification, we conduct a comparative analysis of the prediction performance between SToRI and TaskRes~\citep{yu2023task}.
{TaskRes is a recent method for few-shot image classification, which trains class-specific residual embedding $x_c$ added to initial text embedding $t_c$ to create new classifier $t_c+\alpha x_c$ for each class $c$. Here, $t_c$ denotes the text embedding derived from a given text prompt for class $c$, and $\alpha$ is a hyperparameter for scaling. $x_c$ is trained with cross-entropy loss (refer to Eq.~\eqref{eq:loss_ce_preliminary}).} Such residual embeddings exist in uninterpretable space, rendering the final classifier also uninterpretable. In contrast, SToRI trains only weights, indicating the degree to which each semantic element within a given sentence should be emphasized, thus maintaining interpretability.

Ensuring interpretability, SToRI achieves performance comparable to TaskRes, as presented in Table~\ref{tab:zeroshot_perf}. 
``Base'' refers to custom text prompts including class names, which are generally used in few-shot image classification tasks with CLIP~\citep{yu2023task}. We use both base and CuPL text prompts, with weights trained exclusively on CuPL.
In the 1/2-shot setting, SToRI generally outperforms TaskRes across most datasets. In the 4/8/16-shot setting, it exhibits only a marginal difference, achieving nearly similar performance. This indicates that SToRI provides substantial flexibility to text embeddings, enabling it to be an enhanced text classifier that effectively represents image data. Please refer to Appendix~\ref{sec:appendix_exp_fewshot} for the MetaCLIP results, which align closely with those from CLIP.

\subsection{Retrieval with User-driven Control}\label{sec:exp_user_driven_control}
To assess the effectiveness of SToRI in emphasizing or de-emphasizing specific information based on applied weights, we compare the ordering of retrieved images using text embeddings.

\subsubsection{Experimental Setup}
\paragraph{Dataset}
We use CelebA~\citep{liu2015faceattributes} and CUB~\citep{wah2011caltech} datasets.
The CelebA dataset contains over 200K face images, each annotated with 40 attributes. The CUB dataset contains over 11K bird images, which are annotated with 312 attributes.
Three attributes are chosen to create eight categories based on their presence or absence.
{For the CelebA dataset, each category comprises 100 randomly selected images, resulting in a total of 800 images. For the CUB dataset, all images are used.}
For more details, please refer to Appendix~\ref{sec:appendix_exp_detail_user}.

\paragraph{Image Retrieval with Preference}
We construct a text prompt containing the selected attributes. For instance, the text prompt becomes \texttt{`a photo of a woman with blonde hair, wearing eyeglasses'} for the attributes \textit{female}, \textit{blonde hair}, and \textit{eyeglasses}. Using the text prompt and attribute weights, we obtain a corresponding text embedding through SToRI, followed by sorting the images in descending order of similarity between their image embeddings and the text embedding.

\paragraph{Model} {Most experiments are conducted using CLIP ViT-L/14~\cite{radford2021learning}, unless otherwise specified. Experiments are also conducted using various VLMs, including OpenCLIP~\cite{cherti2023reproducible} and MetaCLIP~\citep{xu2023demystifying}. Reweighting is applied from the 7th block.}

\input{tabs/fig_result_retrieval_density}
\input{tabs/fig_result_retrieval}

\subsubsection{Metric for Preference Retrieval}
Our primary focus is on observing how adjusting weights for specific semantic elements affects the image retrieval order. To facilitate this comparison, we report the average precision score (AP) and precision at rank $k$ (P$_k$) for images with the attributes influenced by the adjusted weights. For instance, when we modify the weight on \texttt{`eyeglasses'}, we consider images with eyeglasses as positive samples and calculate AP and P$_k$.

Additionally, we introduce a novel metric to quantify priority in preference retrieval. We generate a line plot illustrating the proportion of images retrieved for each attribute combination up to the $n$-th retrieved image (see the second row in Figure~\ref{fig:res_retrieval_density}), and calculate the Area Under the Curve (AUC) for each plotted curve. A higher AUC value suggests a faster retrieval of associated visual attribute set, indicating a higher priority in the retrieval process.

\input{tabs/tab_retrieval}

\subsubsection{Results}\label{sec:exp_retrieval_res}
\paragraph{Effect of Emphasizing and De-emphasizing}
Initially, we select three attributes, \textit{female}, \textit{blonde hair}, and \textit{eyeglasses}, and observe the ordering of image retrieval as shown Figure~\ref{fig:res_retrieval_density}.
With the plain text embedding, the initial bin predominantly contains images featuring all selected attributes, followed by a prevalence of images from the `female, no blonde hair, eyeglasses' category. 
When the weight on \texttt{`with blonde hair'} increases and on \texttt{`wearing eyeglasses'} decreases, images belonging to `female, blonde hair, no eyeglasses' are retrieved more prominently.
This suggests that the `blonde hair' gains more representation in the text embedding through reweighting.
The groups with two or more mismatched attributes still rank lower, indicating that our method preserves the meanings of the original text while appropriately reflecting the intention of emphasis and de-emphasis.

We conduct quantitative validation across various text prompts. Table~\ref{tab:retrieval_perf} presents AP and P$_{400}$ scores while controlling weights on attributes. We generate image pools and text prompts from three selected attributes.
The reported scores are based on adjusting the weight for one specific attribute, considering the images containing that attribute as positive samples. Various combinations of attributes, totaling {20 text prompts for the CelebA dataset and 58 text prompts for the CUB dataset}, are used to obtain scores, and their averages and standard deviations are reported. Further details are in Appendix~\ref{sec:appendix_exp_detail_user}.
The results show that modifying the weight of tokens corresponding to a specific attribute in the text prompt results in faster retrieval of images with that attribute (both scores become higher) when the weight increases and slower retrieval when decreases (both scores become lower). This shows that adjusting the weight influences the creation of text embeddings, effectively highlighting or downplaying the corresponding attribute.
{Additional results on more complex scenarios, including those with MetaCLIP, are in Appendix~\ref{sec:appendix_exp_pref}.}

\paragraph{Effect of Weight Control}
{Figure~\ref{fig:res_retrieval_lineplot}} demonstrates the effects of weight control on the AUC scores for the retrieval of each category. As the weight assigned to the \texttt{`with blonde hair'} increases and the weight for \texttt{`wearing eyeglasses'} decreases, there is a noticeable rise in the AUC scores for the two categories that have blonde hair but no eyeglasses. In contrast, categories characterized by the absence of blonde hair and the presence of eyeglasses see a reduction in their AUC scores. When the weight assigned to \texttt{`with blonde hair'} is set to zero, the differentiation between the `female, blonde hair, eyeglasses' and 'female, no blonde hair, eyeglasses' categories is effectively eliminated, resulting in remarkably similar AUC scores. The effect of weight control is consistent across different CLIP models, such as CLIP ViT-B/16, CLIP ViT-L/14, OpenCLIP~\cite{cherti2023reproducible}, and MetaCLIP~\citep{xu2023demystifying}.
This shows that SToRI enables the emphasis or de-emphasis of specific semantics within a text when constructing text embeddings across various models, showcasing its versatility.

%% file: tabs/tab_fewshot_textdescription.tex
\begin{table}
\setlength{\tabcolsep}{5pt}
\centering
\begin{adjustbox}{max width=\linewidth}
\begin{tabular}{lcc}
\toprule
Text & Caltech101 & SUN397 \\ \midrule
CuPL & 97.42$\pm$0.23
 & 79.54$\pm$0.12
\\
CuPL+Nonsensical tokens & 97.30$\pm$0.15
 & 79.11$\pm$0.10
\\ \bottomrule
\end{tabular}
\end{adjustbox}
\caption{Accuracy (\%) on 16-shot image classification.
}
\label{tab:compare_text_few_perf}
\end{table}

%% file: tabs/tab_fewshot.tex
\begin{table*}
\centering
\begingroup
\begin{adjustbox}{max width = 1.0\textwidth}
\begin{tabular}{lll|cccccc|c}
\toprule
 & Method & Text & ImageNet & DTD & Flowers102 & SUN397 & Caltech101 & Food101 & AVG \\ \midrule
\multirow{3}{*}{1shot} & TaskRes & Base & {75.95}$\pm$0.03 & 55.40$\pm$0.27 & 81.16$\pm$0.44 & 68.10$\pm$0.16 & 94.28$\pm$0.11 & 90.30$\pm$0.10 & 77.53 \\
 & TaskRes & Base+CuPL & 74.69$\pm$0.04 & {65.66}$\pm$0.82 & {90.07}$\pm$0.79 & {73.52}$\pm$0.49 & {95.89}$\pm$0.57 & {90.35}$\pm$0.36 & \underline{81.70} \\
 & SToRI (Ours) & Base+CuPL & {76.68}$\pm$0.15 & {65.82}$\pm$0.98 & {89.05}$\pm$0.58 & {72.88}$\pm$0.20 & {96.27}$\pm$0.67 & {91.34}$\pm$0.12 & \textbf{82.01} \\ \midrule
\multirow{3}{*}{2shot} & TaskRes & Base & 76.03$\pm$0.00 & 55.52$\pm$0.48 & 81.50$\pm$0.62 & 69.53$\pm$0.14 & 94.54$\pm$0.05 & 90.49$\pm$0.05 & 77.93 \\
 & TaskRes & Base+CuPL & 75.55$\pm$0.04 & {66.45}$\pm$1.57 & {92.38}$\pm$0.69 & 75.69$\pm$0.29 & 96.96$\pm$0.27 & 90.64$\pm$0.38 & \underline{82.95} \\
 & SToRI (Ours) & Base+CuPL & {77.36}$\pm$0.23 & 66.37$\pm$1.01 & 91.56$\pm$0.60 & {75.75}$\pm$0.04 & {97.15}$\pm$0.13 & {91.49}$\pm$0.24 & \textbf{83.28} \\ \midrule
\multirow{3}{*}{4shot} & TaskRes & Base & 76.16$\pm$0.02 & 55.85$\pm$0.12 & 81.65$\pm$0.28 & 71.15$\pm$0.09 & 94.58$\pm$0.09 & 90.44$\pm$0.05 & 78.31 \\
 & TaskRes & Base+CuPL & {76.42}$\pm$0.03 & {70.76}$\pm$1.12 & {93.22}$\pm$0.37 & {77.20}$\pm$0.08 & {97.40}$\pm$0.21 & 91.45$\pm$0.15 & \textbf{84.41} \\
 & SToRI (Ours) & Base+CuPL & {77.90}$\pm$0.05 & 69.03$\pm$1.48 & 92.46$\pm$0.09 & 76.89$\pm$0.02 & 97.39$\pm$0.08 & {91.68}$\pm$0.07 & \underline{84.22} \\ \midrule
\multirow{3}{*}{8shot} & TaskRes & Base & 76.87$\pm$0.05 & 58.14$\pm$0.07 & 86.82$\pm$0.19 & 74.52$\pm$0.07 & 96.17$\pm$0.08 & 91.12$\pm$0.07 & 80.60 \\
 & TaskRes & Base+CuPL & 77.97$\pm$0.02 & {73.42}$\pm$0.86 & {98.17}$\pm$0.25 & {77.54}$\pm$0.16 & {97.00}$\pm$0.28 & {91.27}$\pm$0.11 & \textbf{85.89} \\
 & SToRI (Ours) & Base+CuPL & {78.38}$\pm$0.13 & 72.03$\pm$0.60 & 97.51$\pm$0.43 & {78.34}$\pm$0.13 & {96.98}$\pm$0.29 & 90.50$\pm$0.05 & \underline{85.62} \\ \midrule
\multirow{3}{*}{16shot} & TaskRes & Base & 77.34$\pm$0.03 & 61.47$\pm$0.16 & 90.85$\pm$0.21 & 76.01$\pm$0.24 & 96.75$\pm$0.07 & 91.30$\pm$0.10 & 82.29 \\
 & TaskRes & Base+CuPL & {79.18}$\pm$0.10 & {77.05}$\pm$0.65 & {99.07}$\pm$0.11 & {78.98}$\pm$0.10 & {97.65}$\pm$0.23 & {91.49}$\pm$0.08 & \textbf{87.24} \\
 & SToRI (Ours) & Base+CuPL & 79.03$\pm$0.13 & 74.94$\pm$0.10 & {98.55}$\pm$0.23 & {79.61}$\pm$0.11 & {97.43}$\pm$0.20 & 91.18$\pm$0.10 & \underline{86.79} \\
 \bottomrule
\end{tabular}
\end{adjustbox}
\endgroup
\caption{Accuracy (\%) on few-shot classification with CLIP ViT-L/14. The results include mean values with standard deviation across three runs. The results of TaskRes are reproduced. The best performance is indicated in bold, while the second-best performance is underlined.}
\label{tab:zeroshot_perf}
\end{table*}

%% file: tabs/fig_result_retrieval_density.tex
\begin{figure}[t]
\begin{center}
\centerline{\includegraphics[width=\linewidth]{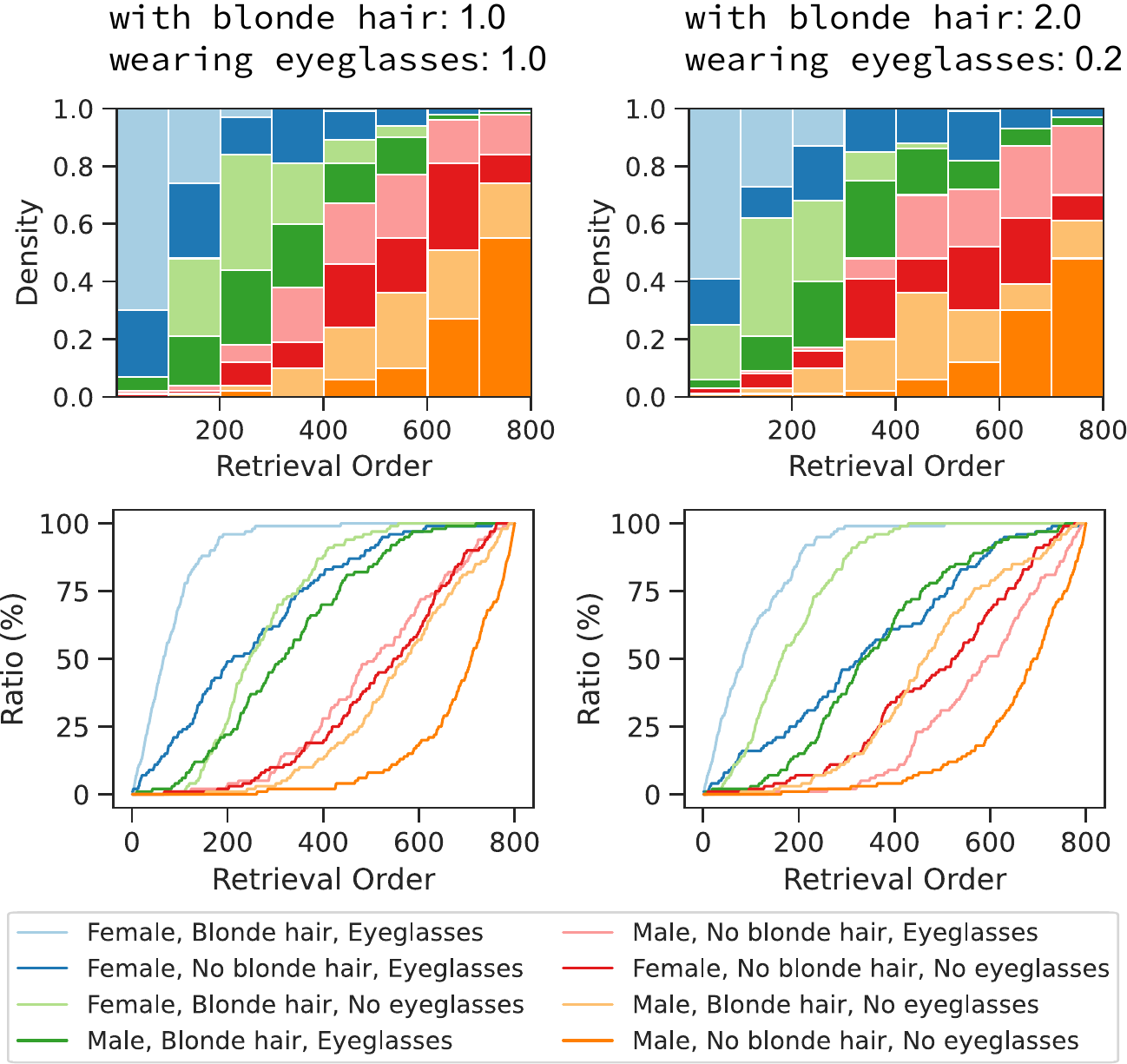}}
\caption{Results of preference retrieval using the text prompt \texttt{`a photo of a woman with blonde hair, wearing eyeglasses'}. The first row shows density plots with the retrieval order, and the second row visualizes the ratio of retrieved samples within each category. The left column shows results from a plain text prompt, whereas the right column depicts the results when the weights are adjusted. Best viewed in color.}
\label{fig:res_retrieval_density}
\end{center}
\end{figure}

%% file: tabs/fig_result_retrieval.tex
\begin{figure*}[htb!]
\begin{center}
\centerline{\includegraphics[width=\linewidth]{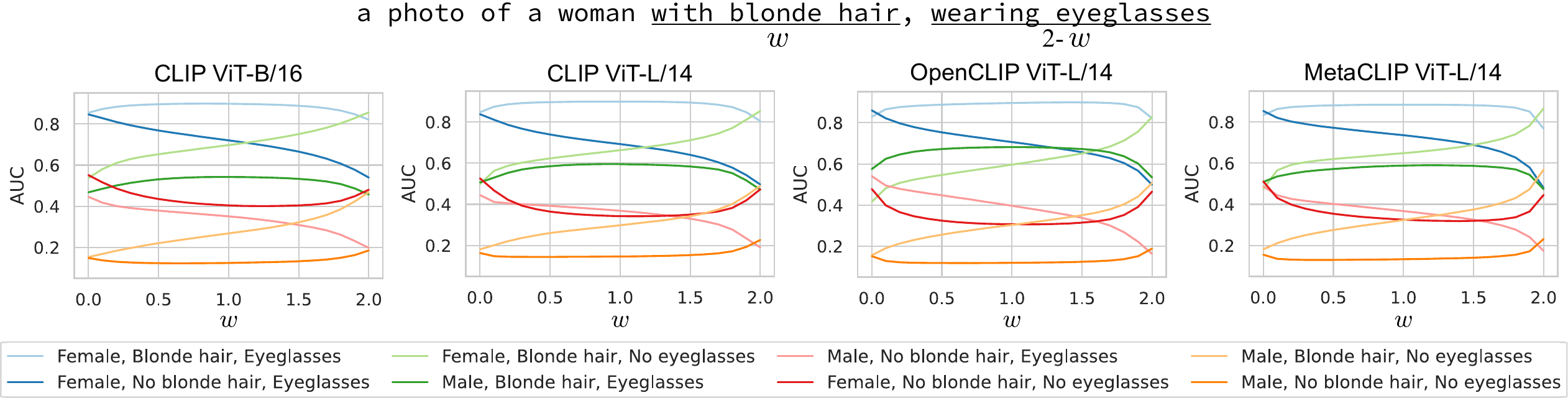}}
\caption{AUC scores from preference retrieval with varying weights. The text prompt is \texttt{`a photo of a woman with blonde hair, wearing eyeglasses'}. The weights on \texttt{`with blonde hair'} and \texttt{`wearing eyeglasses'} are $w$ and $(2-w)$, respectively, which are adjusted simultaneously in opposite direction. Best viewed in color.}
\label{fig:res_retrieval_lineplot}
\end{center}
\end{figure*}

%% file: tabs/tab_retrieval.tex
\begin{table}[t]
\setlength{\tabcolsep}{2.8pt}
\centering
\begin{adjustbox}{max width = \linewidth}
\begin{tabular}{cccc}
\toprule
& \multicolumn{2}{c}{CelebA} & CUB \\
 & AP & P$_{400}$ & AP \\ \midrule
Plain ($w=1.0$)  & 0.752$\pm$0.089 & {0.679$\pm$0.084} & {0.154$\pm$0.070}\\ \midrule
Emphasized & 0.773$\pm$0.084 & 0.697$\pm$0.068 & 0.183$\pm$0.079\\
~($w=1.5$)& \small{\textbf{$\Delta$0.021$\pm$0.011}} & \small{\textbf{$\Delta$0.017$\pm$0.009}} & \small{\textbf{$\Delta$0.029$\pm$0.018}}\\ \midrule
De-emphasized & 0.709$\pm$0.096 & 0.648$\pm$0.072 & 0.116$\pm$0.057\\ 
~($w=0.5$)& \small{\textbf{$\Delta$-0.043$\pm$0.021}} & \small{\textbf{$\Delta$-0.031$\pm$0.031}} & \small{\textbf{$\Delta$-0.038$\pm$0.021}}\\
\bottomrule
\end{tabular}
\end{adjustbox}
\caption{Retrieval performance on attributes of the CelebA and CUB datasets with CLIP ViT-L/14. The results show mean values with standard deviation across multiple controlled attributes.}
\label{tab:retrieval_perf}
\end{table}

%% file: texts/02.related_work.tex
\section{Related Works}\label{sec:related}

\paragraph{VLMs and Interpretability}
In recent vision tasks, interpretative analysis in natural language becomes popular rather than relying solely on visual form. VLMs like CLIP have commonly been employed to connect the image and text feature spaces for explanations.
\citet{kim2023grounding} utilized CLIP to derive concept activation vector~\citep{kim2018interpretability} in vision model, while \citet{yuksekgonul2023posthoc} and \citet{oikarinen2023labelfree} leveraged CLIP to identify whether concepts defined in text are present in images.
\citet{menon2023visual} and \citet{pratt2023does} crafted text prompts to explain image classes using large language models (LLMs), applying them for zero-shot classification with CLIP.
Beyond simply utilizing CLIP’s shared embedding space, several works have focused on enhancing the interpretability of this space. \citet{chen-etal-2023-stair} introduced the STAIR model, which improves interpretability by fine-tuning CLIP to generate sparse, more interpretable representations. \citet{bhalla2024interpreting}, on the other hand, proposed decomposing image embeddings into conceptual text embeddings, combining concepts within the embedding space for clearer explanations.
In contrast to these methods, we ensure interpretability by manipulating token weighting within a given description’s context, without the need for model fine-tuning or embedding decomposition. This approach allows for practical applications in controlling interpretation, such as adjusting token importance to reflect user preferences or contextual emphasis, while also enabling detailed analysis of interpretability.

\paragraph{Few-shot Image Classification}
CLIP exhibits promising performance in image recognition tasks, leading to the development of various few-shot learning approaches. CoOp~\citep{zhou2022learning} and CoCoOp~\citep{zhou2022conditional} are representative methods based on prompt tuning. Tip-Adapter~\citep{zhang2022tip} integrates an extra adapter unit following the encoders. TaskRes~\citep{yu2023task} involves training task-specific residual text embeddings for each category.
While these approaches incorporate extra trainable parameters outside an interpretable framework and thus do not guarantee interpretability, our framework enables the training of classifiers while ensuring interpretability.

\paragraph{Enrich Textual Representation}
In text-to-image generation, several approaches have been developed to enrich textual representation. Prompt weighting\footnote{https://huggingface.co/docs/diffusers/using-diffusers/weighted\_prompts} is a common technique in Stable Diffusion~\citep{rombach2022high}, which multiplies weights to individual output token embeddings prior to supplying them to the image generation model. Prompt-to-Prompt controls cross-attention between noise images and text embeddings~\citep{hertz2022prompt}. Additionally, \citet{ge2023expressive} proposed a richer text editor that allows users to define various input conditions for image generation, such as coloring and footnotes.
A similar approach has been explored in text generation. \citet{zhang2024tell} introduced a method that enables large language models to process text with user-defined emphasis by reducing attention to unspecified parts of the text.
\citet{zhang2024prompt} proposed Prompt Highlighter, which highlights tokens during generation process with Multi-Modal LLMs.
While prior works have focused on image and text generation, typically using only user-defined attention, our work innovates by developing enriched textual representations for image recognition and proposing an approach for deriving these representations from data. This distinctive approach establishes a new avenue for incorporating linguistic context in visual understanding.

%% file: texts/05.conclusion.tex
\section{Conclusion}\label{sec:conclusion}
We propose SToRI, a framework that builds interpretable text embeddings by reweighting semantic tokens in CLIP. This approach innovatively enhances the explanatory power of natural language in vision tasks. Our control strategies enable tuning of text embeddings for classification and retrieval while maintaining interpretability. SToRI can be easily applied to any model based on attention mechanisms and has potential scalability across various vision tasks. Additionally, our method of reweighting text prompt tokens during text encoding can similarly be applied to image encoding. This approach can allow for the emphasis of specific image regions. The extension to multi-modal tasks using diverse VLMs remains a topic for future work.

%% file: texts/06.limitation.tex
\section{Limitations}\label{sec:limitation}
Our method is focusing on controlling the attention of each semantic element within a given natural language sentence, rather than generating new textual information.
Therefore, one of the limitations of our method is its dependence on the richness and quality of the given texts. For example, when using data to train a classifier, if the given text lacks sufficient rich information, adjusting the attention may not sufficiently enlarge the text embedding space. This difficulty in expanding the embedding space makes it challenging to establish a basis for improving classification performance and explaining data.

Additionally, we do not consider the inherent black box characteristics of CLIP. However, if this model has undergone sufficient testing and is deemed reliable, the advantage of our method lies in additional optimization and control being in a reliable and controllable space.

\section{Ethics Statement}
Our goal is to employ contollability when building text embeddings. This enables for users to emphasize or deemphasize a certain part of textual information and improving text embeddings for vision tasks, ensuring interpretability. We believe this work can be used to build trustful AI systems by providing natural language interpretation.

If CLIP in use is biased towards the attributes targeted for reweighting, it may also affect other related attributes. The best approach to address this issue is to use CLIP that has been trained to reduce bias. However, if a biased CLIP must be used, designing text prompts that can help mitigate the bias could be a potential strategy to consider.

%% file: texts/appendix.tex
\appendix

\section{Experimental Details}\label{sec:appendix_exp_detail}
\subsection{Few-shot Image Classification}\label{sec:appendix_exp_detail_data}
We use Adam optimizer with the cosine learning rate scheduler~\cite{loshchilov2017sgdr} following the training scheme of TaskRes~\citep{yu2023task}. For CLIP, the learning rate is set to $1\times 10^{-2}$ for the ImageNet and SUN397 datasets, $0.1$ for the Food101 dataset and for 8/16-shot scenarios on the DTD and Flower102 datasets, and $5\times 10^{-2}$ for the other datasets.
{For MetaCLIP, the learning rate is set to $1\times 10^{-2}$ for the ImageNet and SUN397 datasets, $0.1$ for Flower102 dataset, and $5\times 10^{-2}$ for the other datasets. The weight decay is set to $0$ for both models.}
When reproducing TaskRes, the learning rate is set to $2\times 10^{-5}$ for the ImageNet dataset and $2\times 10^{-4}$ for the other datasets. The weight decay is set to $0.005$ and $\alpha$ is set to $0.5$. The training is conducted with a batch size of 256.
All experiments are implemented using PyTorch~\cite{paszke2017automatic}, and we use official code base released by \citet{yu2023task} to reproduce TaskRes.

\subsection{Image Retrieval}\label{sec:appendix_exp_detail_user}
\textbf{CelebA.~}
We initially select 11 attributes with a zero-shot classification performance of AUROC 0.75 or higher with CLIP on test set. For zero-shot classification, we create text prompt for each attribute and calculate AUROC using the similarity between the test set images and the text prompt. For example, when evaluating the attribute \textit{smiling}, we use the text prompt \texttt{`a photo of a smiling person'}. Among the identified 11 attributes, we create combinations of three and five attributes, each including either \textit{female} or \textit{male}. For the combinations of three attributes, we filter out the combinations where all eight categories contain fewer than 100 images. We conduct image retrieval with total 20 numbers of text prompts based on the combinations of attributes, as shown in Table~\ref{tab:text_prompt_celeba}. Details on combinations of five attributes can be found in Appendix~\ref{sec:appendix_exp_pref}.

\noindent\textbf{CUB.~}
{Following the filtering process described by \citet{koh2020concept}, we initially retain 112 attributes. We then select 15 attributes that achieve a zero-shot classification performance with AUROC 0.75 or higher using CLIP. Notably, the attribute labels in the CUB dataset are finely detailed and related to various parts of birds, which poses a challenge for CLIP in differentiation. With the chosen attributes, we form combinations of three attributes that do not share the same color, yielding 58 combinations. The text prompt we use is \texttt{`a photo of a bird, which has [text for attribute1], has [text for attribute2], and has [text for attribute3]'}. Table~\ref{tab:text_prompt_cub} presents 15 attributes and their corresponding texts.}

We use all the datasets and models solely for academic research purposes and do not employ them for improper intentions.

\section{Metric for Preference Retrieval}
To quantify priority in preference retrieval, we introduce a novel metric using the area under the curve (AUC). First, we obtain the top $n$ images with the highest similarity to the text embedding. We then calculate the proportion of images from each category that fall within rank $n$ and plot these proportions as a function of $n$, as shown in the second row of Figure~\ref{fig:res_retrieval_density}. The AUC of these plots represents how quickly images from each category are retrieved, providing a measure of retrieval efficiency for each category.

\section{Additional Experimental Results}\label{sec:appendix_exp_res}

\subsection{{Comparison to Prompt Weighting}}
We compare SToRI with prompt weighting, a technique often used in text-to-image generation via Stable Diffusion~\citep{rombach2022high}. Prompt weighting multiplies weights by the difference in output token embeddings when provided with a text prompt versus an empty one. Unlike Stable Diffusion, which utilizes all output token embeddings, we aim to build a vector form of text embedding from \texttt{[EOS]} token. {Therefore, we modify prompt weighting for use at an intermediate layer, which we refer to as modified prompt weighting, and compare it with SToRI on preference-based image retrieval.}

\input{tabs/fig_result_retrieval_promptweighting}
As depicted in Figure~\ref{fig:res_retrieval_lineplot_pw}(a), the modified prompt weighting influences the significance of tokens similarity to SToRI. However, the change in AUC is not gradual; it remains nearly static when weights fall below 0.5 or above 1.5. As shown in Figure~\ref{fig:res_retrieval_lineplot_pw}(b), even when the weight for \texttt{`with blonde hair'} increases significantly, SToRI consistently raises the AUC for the category `female, blonde hair, no eyeglasses'. In contrast, the AUC with modified prompt weighting initially increases but subsequently decreases, indicating augmented weight fails to heighten emphasis.
This could stem from the scaling of intermediate embeddings which, when overextended, surpasses the scale that the text encoder is pre-trained to deal with, lessening the intended effect of emphasis. SToRI, on the other hand, adjusts normalized attention scores within the self-attention mechanism, ensuring that as weight escalates, the relevant tokens consistently obtain attention scores approaching 1, thus preserving the desired impact.

\subsection{{Additional Results for Few-shot Classification}}\label{sec:appendix_exp_fewshot}
{Table~\ref{tab:zeroshot_perf_metaclip} compares few-shot classification performances of SToRI and TaskRes\mbox{~\cite{yu2023task}} on MetaCLIP ViT-L/14. Similar to the results on CLIP, the results show that SToRI achieves performance comparable to TaskRes, which uses uninterpretable classifiers. These experiments further support our findings, demonstrating our approach's effectiveness across models and highlighting its adaptability and scalability.}

\subsection{Additional Examples for Interpretation}
Figures~\ref{fig:result_interpretation_imagenet_app} and \ref{fig:result_interpretation_dtd_app} present examples of text prompts and the corresponding trained weights for each token within the ImageNet and DTD datasets, respectively. Higher weights are assigned to word tokens that effectively represent images.

\subsection{{Additional Results for Retrieval}}\label{sec:appendix_exp_pref}
\input{tabs/tab_retrieval_app}
We assess SToRI in the context of preference-based retrieval by assigning different weights to multiple attributes to explore how varying weight magnitudes affect emphasis. We create combinations of three attributes and assign them different weights: one attribute is assigned a weight of 2.0, another a weight of 1.5, and the remaining one a weight of 1.0. We then compare the retrieval performance for attributes with weights of 1.5 and 2.0.
Table~\ref{tab:retrieval_perf_complex} demonstrates that the retrieval performance of the attribute with a weight of 1.5 increases, while the attribute with a weight of 2.0 shows an even greater increase in retrieval performance. This indicates that when semantic tokens are assigned different weights, the emphasis effect increases proportionally with the assigned weights compared to plain text. This highlights the significance of the magnitude of weights.

Table~\ref{tab:retrieval_perf_metaclip} presents the results on MetaCLIP ViT-L/14 when adjusting the weight of one attribute among three within combinations of three attributes (as outlined in Section~\ref{sec:exp_user_driven_control}). The results demonstrate that emphasizing or de-emphasizing an attribute in MetaCLIP leads to increased or decreased retrieval performance for images with the specified attribute, showcasing the scalability of SToRI across models.

To evaluate SToRI in more complex attribute combinations, we perform retrieval using combinations of five attributes. Only the following five attributes result in images for all 32 possible categories formed by combinations of the five attributes: \textit{male} or \textit{female}, \textit{smiling}, \textit{bangs}, \textit{gray hair}, and \textit{eyeglasses}. We use two text prompts for \textit{male} and \textit{female}. We randomly select five images for each category, resulting in a total of 160 images.
Table~\ref{tab:retrieval_perf_5attr} presents the results on CLIP and MetaCLIP ViT-L/14 when adjusting the weight of one attribute among five. These findings underscore a consistent trend of increasing retrieval scores when attributes are emphasized and decreasing scores when attributes are de-emphasized, across different attribute combinations.

\subsection{Computational Cost}
We calculate runtime for applying SToRI compared to plain text embeddings, as reported in Table~\ref{tab:computation_cost}. The experiment is done on RTX A5000 and the reported values are mean values from 28K runs. Since SToRI only multiplies predefined weights when calculating attention scores, the runtime is not significantly different from that of plain text embeddings.

\input{tabs/tab_cost}

\subsection{Position for Reweighting}\label{sec:appendix_exp_position}
Figure~\ref{fig:ablation_layer_retrieval}(a) compares the changes in AUC scores when we start reweighting at various positions. The reweighting process is applied to all blocks following a specific block. There is not a significant difference when we initiate token reweighting at intermediate positions. However, when token reweighting is applied to all blocks (from 1st block), a sharp bend is observed at 0.1 when the weight decreases. This is unlike other cases, which show a smooth decrease or increase in all scenarios. It is presumed that this abrupt occurrence is due to tokens in the specified position being completely disregarded when the weight becomes 0, leading to sudden gaps in those areas.

Figure~\ref{fig:ablation_layer_retrieval}(b) illustrates that when reweighting is applied only within a single specific intermediate block, the effects of emphasis or de-emphasis are scarcely observed.
This suggests that if reweighting is confined within a single intermediate block, its effects in the subsequent blocks are counteracted, indicating that it should be applied in the subsequent blocks to emphasize or de-emphasize semantic tokens.

Figure~\ref{fig:ablation_layer_fewshot} shows the changes in few-shot classification performance when we start reweighting at various positions. The reweighting process is applied to all blocks following a specific block. Like the results in image retrieval, there is not a significant difference when we initiate token reweighting at intermediate positions.

\subsection{Studies on Failure Cases}

There are some cases where non-semantic elements are assigned high weights in differentiating classes, which may appear illogical to a human observer.
For example, in Figure~\ref{fig:result_interpretation_imagenet_app}, `.' is assigned a high weight. This occurrence likely results from the training process, where it's advantageous to emphasize not only the semantic meaning but also to differentiate from other classes. Hence, `.' is not emphasized for other classes but is for this specific class. This can also be observed in Figure~\ref{fig:result_interpretation_cub}, where in the comparison of Blue headed Vireo vs. Warbling Vireo, `bird' is emphasized only for Blue headed Vireo, and `reo' is more emphasized only for Warbling Vireo.

\input{tabs/fig_appendix_interpretation}
\input{tabs/tab_fewshot_metaclip}
\input{tabs/tab_celeba_textprompt}
\input{tabs/fig_ablation_layer}
\input{tabs/fig_ablation_layer_fewshot}

%% file: tabs/fig_result_retrieval_promptweighting.tex
\begin{figure}[t]
\begin{center}
\centerline{\includegraphics[width=\linewidth]{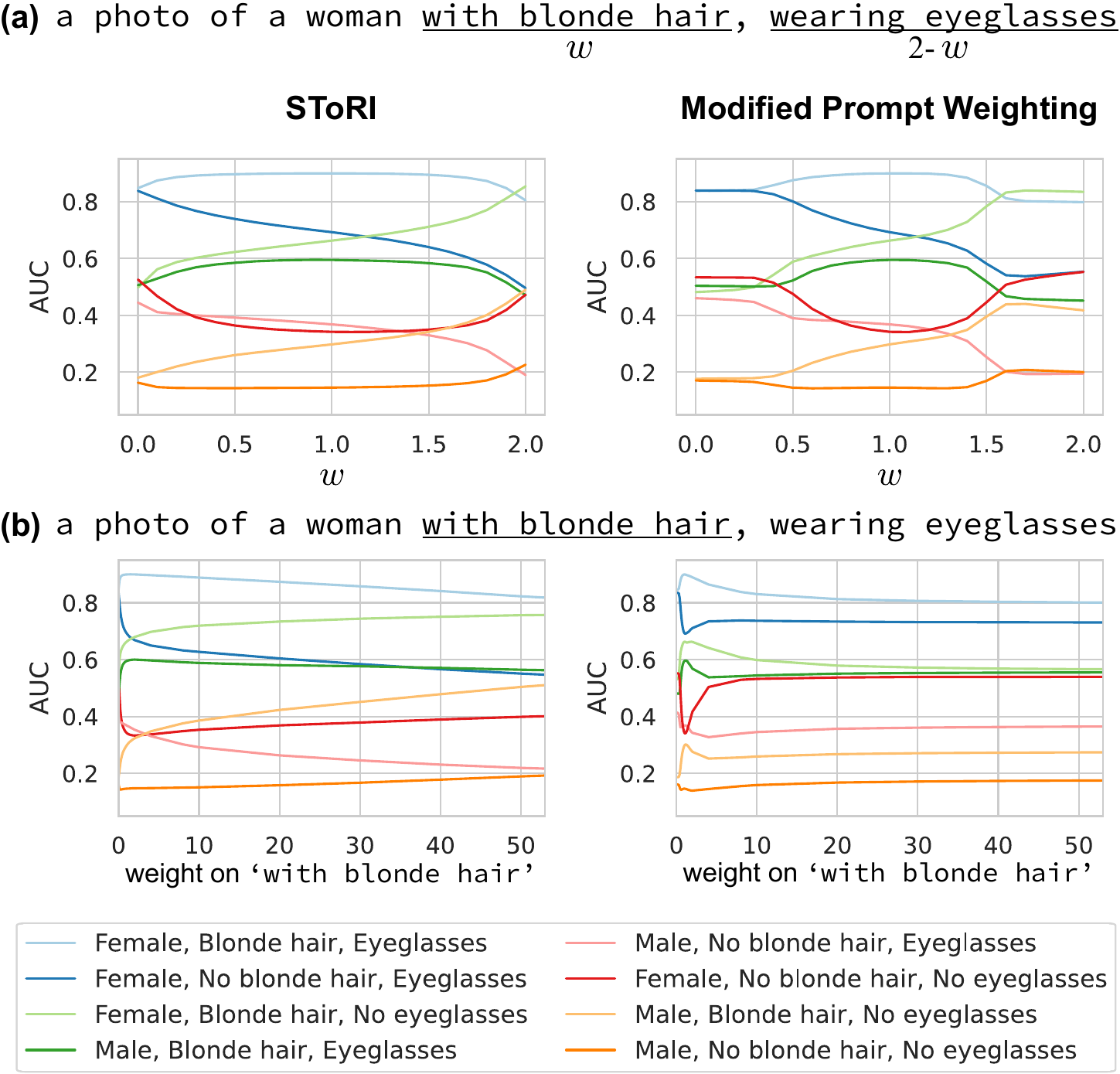}}
\caption{AUC scores from preference retrieval with varying weights. The text prompt is \texttt{`a photo of a woman with blonde hair, wearing eyeglasses'}. (a) The weights on \texttt{`with blonde hair'} and \texttt{`wearing eyeglasses'} are $w$ and $(2-w)$, respectively, which are adjusted simultaneously in opposite direction. (b) Only the weight on \texttt{`with blonde hair'} is adjusted. Best viewed in color.}
\label{fig:res_retrieval_lineplot_pw}
\end{center}
\end{figure}

%% file: tabs/tab_retrieval_app.tex
\begin{table}[t]
\setlength{\tabcolsep}{3pt}
\centering
\begin{adjustbox}{max width = \linewidth}
\begin{tabular}{llcc}
\toprule
 & & AP & P$_{400}$ \\ \midrule
\multicolumn{2}{l}{Plain ($w=1.0$)} & 0.752$\pm$0.089 & 0.679$\pm$0.070 \\ \midrule
\multirow{4}{*}{Emphasized} & Attribute & 0.754$\pm$0.085 & 0.681$\pm$0.064 \\ 
& with $w=1.5$ & \small{\textbf{$\Delta$0.003$\pm$0.017}} & \small{\textbf{$\Delta$0.002$\pm$0.016}} \\ \cmidrule{2-4}
& Attribute  & 0.776$\pm$0.082 & 0.698$\pm$0.064 \\
& with $w=2.0$ & \small{\textbf{$\Delta$0.024$\pm$0.019}} & \small{\textbf{$\Delta$0.019$\pm$0.016}} \\
\bottomrule
\end{tabular}
\end{adjustbox}
\caption{{Retrieval performance on attributes of the CelebA dataset when two attributes are assigned different weights. The results show mean values with standard deviation across multiple controlled attributes.}}
\label{tab:retrieval_perf_complex}
\end{table}

\begin{table}[t]
\setlength{\tabcolsep}{3pt}
\centering
\begin{adjustbox}{max width = \linewidth}
\begin{tabular}{cccc}
\toprule
& \multicolumn{2}{c}{CelebA} & CUB \\
 & AP & P$_{400}$ & AP \\ \midrule
Plain ($w=1.0$)  & {0.753$\pm$0.088} & {0.681$\pm$0.062} & {0.148$\pm$0.055}\\ \midrule
Emphasized & 0.774$\pm$0.086 & 0.699$\pm$0.063 & 0.195$\pm$0.074\\
~($w=1.5$)& \small{\textbf{$\Delta$0.021$\pm$0.011}} & \small{\textbf{$\Delta$0.018$\pm$0.009}} & \small{\textbf{$\Delta$0.047$\pm$0.026}}\\ \midrule
De-emphasized & 0.709$\pm$0.087 & 0.647$\pm$0.057 & 0.098$\pm$0.035\\ 
~($w=0.5$)& \small{\textbf{$\Delta$-0.044$\pm$0.022}} & \small{\textbf{$\Delta$-0.035$\pm$0.016}} & \small{\textbf{$\Delta$-0.051$\pm$0.026}}\\
\bottomrule
\end{tabular}
\end{adjustbox}
\caption{{Retrieval performance on attributes of the CelebA and CUB datasets with MetaCLIP ViT-L/14. The results show mean values with standard deviation across multiple controlled attributes.}}
\label{tab:retrieval_perf_metaclip}
\end{table}

\begin{table}
\setlength{\tabcolsep}{3pt}
\centering
\begin{adjustbox}{max width = \linewidth}
\begin{tabular}{cccc}
\toprule
 & & AP & P$_{80}$ \\ \midrule
\multirow{5}{*}{CLIP} & Plain ($w=1.0$) & 0.684$\pm$0.097& 0.627$\pm$0.062 \\ \cmidrule{2-4}
&Emphasized & 0.705$\pm$0.099 & 0.643$\pm$0.069 \\ 
&($w=1.5$) & \small{\textbf{$\Delta$0.021$\pm$0.009}} & \small{\textbf{$\Delta$0.015$\pm$0.012}} \\ \cmidrule{2-4}
&De-emphasized & 0.643$\pm$0.086 & 0.601$\pm$0.054 \\
&($w=0.5$) & \small{\textbf{$\Delta$-0.041$\pm$0.019}} & \small{\textbf{$\Delta$-0.026$\pm$0.012}} \\ \midrule
\multirow{5}{*}{MetaCLIP} & Plain ($w=1.0$) & 0.689$\pm$0.074 & 0.631$\pm$0.062 \\ \cmidrule{2-4}
&Emphasized & 0.713$\pm$0.078 & 0.646$\pm$0.062 \\ 
&($w=1.5$) & \small{\textbf{$\Delta$0.023$\pm$0.008}} & \small{\textbf{$\Delta$0.015$\pm$0.011}} \\ \cmidrule{2-4}
&De-emphasized & 0.644$\pm$0.064 & 0.602$\pm$0.057 \\
&($w=0.5$) & \small{\textbf{$\Delta$-0.045$\pm$0.020}} & \small{\textbf{$\Delta$-0.029$\pm$0.014}} \\
\bottomrule
\end{tabular}
\end{adjustbox}
\caption{{Retrieval performance on the CelebA dataset with CLIP and MetaCLIP ViT-L/14 when five attributes are combined. The results show mean values with standard deviation across multiple controlled attributes.}}
\label{tab:retrieval_perf_5attr}
\end{table}

%% file: tabs/tab_cost.tex
\begin{table}[t]
\centering
\begin{adjustbox}{max width = \linewidth}
\begin{tabular}{lcc}
\toprule
Method & Plain Text Embeddings & SToRI \\ \midrule
Relative Run Time & 1.00 & 1.02\\ 
\bottomrule
\end{tabular}
\end{adjustbox}
\vskip -0.05in
\caption{Relative compuational cost}
\label{tab:computation_cost}
\end{table}

%% file: tabs/fig_appendix_interpretation.tex
\begin{figure*}[t]
\begin{center}
\centerline{\includegraphics[width=\linewidth]{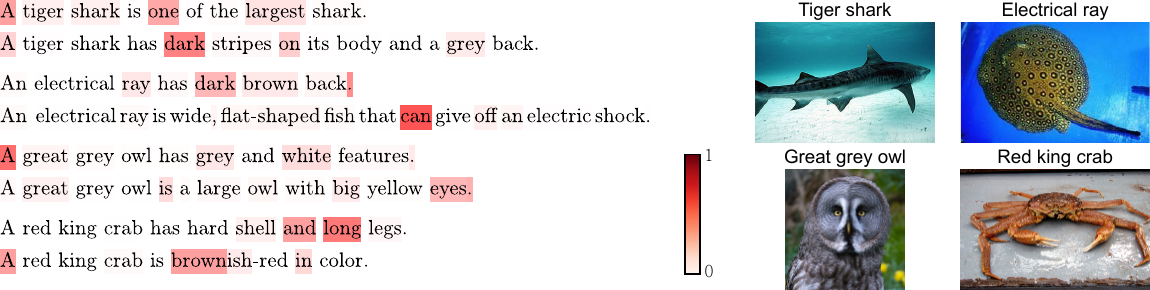}}
\caption{Text prompts and corresponding weights on the ImageNet dataset are provided as examples after training with data. For visualization, the weights are normalized to sum up 1. The figures on the right display an example image for each class.}
\label{fig:result_interpretation_imagenet_app}
\end{center}
\end{figure*}

\begin{figure*}[t]
\begin{center}
\centerline{\includegraphics[width=\linewidth]{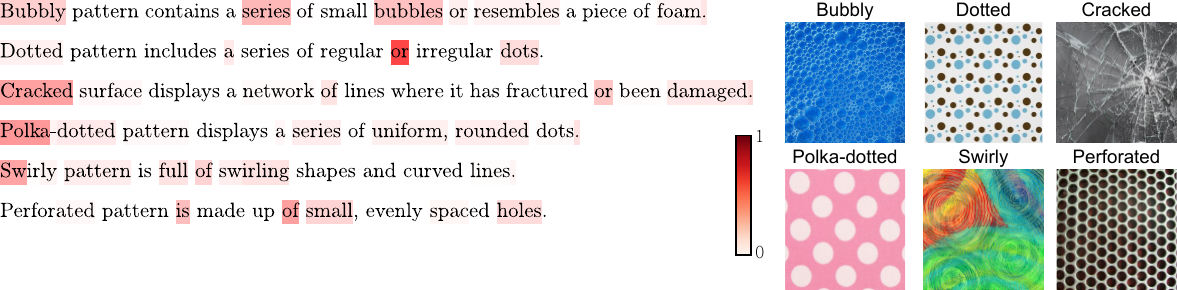}}
\caption{Text prompts and corresponding weights on the DTD dataset are provided as examples after training with data. For visualization, the weights are normalized to sum up 1. The figures on the right display an example image for each class.}
\label{fig:result_interpretation_dtd_app}
\end{center}
\end{figure*}

%% file: tabs/tab_fewshot_metaclip.tex
\begin{table*}
\centering
\begingroup
\begin{adjustbox}{max width = 1.0\textwidth}
\begin{tabular}{lll|cccccc|c}
\toprule
 & Method & Text & ImageNet & DTD & Flowers102 & SUN397 & Caltech101 & Food101 & AVG \\ \midrule
\multirow{3}{*}{1shot} & TaskRes & Base & 79.38$\pm$0.02 & 67.91$\pm$0.26 & 83.75$\pm$0.16 & 74.89$\pm$0.08 & 97.21$\pm$0.15 & 90.63$\pm$0.04 & 82.29  \\
 & TaskRes & Base+CuPL & 79.59$\pm$0.22 & 72.79$\pm$0.54 & 92.26$\pm$0.10 & 76.16$\pm$0.2 & 97.59$\pm$0.19 & 90.28$\pm$0.15 & \textbf{84.78} \\
 & SToRI (Ours) & Base+CuPL & 79.44$\pm$0.17 & 72.66$\pm$0.73 & 92.38$\pm$0.75 & 76.05$\pm$0.38 & 97.46$\pm$0.23 & 90.12$\pm$0.22 & \underline{84.68} \\ \midrule
\multirow{3}{*}{2shot} & TaskRes & Standard & 79.46$\pm$0.01 & 67.93$\pm$0.18 & 84.03$\pm$0.13 & 75.71$\pm$0.13 & 97.48$\pm$0.07 & 90.83$\pm$0.03 & 82.57 \\
 & TaskRes & Base+CuPL & 80.23$\pm$0.14 & 74.27$\pm$1.08 & 94.42$\pm$0.08 & 77.64$\pm$0.28 & 98.20$\pm$0.08 & 90.68$\pm$0.22 & \underline{85.91} \\
 & SToRI (Ours) & Base+CuPL & 79.98$\pm$0.16 & 73.76$\pm$1.38 & 95.09$\pm$0.45 & 78.21$\pm$0.27 & 98.04$\pm$0.02 & 90.57$\pm$0.18 & \textbf{85.94} \\ \midrule
\multirow{3}{*}{4shot} & TaskRes & Standard & 79.58$\pm$0.00 & 68.34$\pm$0.22 & 84.07$\pm$0.12 & 76.66$\pm$0.06 & 97.44$\pm$0.06 & 90.82$\pm$0.02 & 82.82 \\
 & TaskRes & Base+CuPL & 80.68$\pm$0.04 & 76.91$\pm$1.24 & 94.94$\pm$0.18 & 78.88$\pm$0.11 & 98.16$\pm$0.11 & 90.85$\pm$0.07 & \underline{86.74} \\
 & SToRI (Ours) & Base+CuPL & 80.53$\pm$0.09 & 75.91$\pm$0.39 & 96.28$\pm$0.31 & 79.38$\pm$0.14 & 98.01$\pm$0.33 & 90.73$\pm$0.13 & \textbf{86.81} \\ \midrule
\multirow{3}{*}{8shot} & TaskRes & Standard & 80.03$\pm$0.08 & 69.7$\pm$0.45 & 90.12$\pm$0.07 & 78.87$\pm$0.04 & 97.84$\pm$0.10 & 91.30$\pm$0.03& 84.64 \\
 & TaskRes & Base+CuPL & 81.30$\pm$0.12& 78.88$\pm$0.10 & 98.55$\pm$0.17 & 78.87$\pm$0.17 & 98.22$\pm$0.07 & 90.81$\pm$0.18 & \textbf{87.77} \\
 & SToRI (Ours) & Base+CuPL & 81.01$\pm$0.18 & 78.39$\pm$0.27 & 98.04$\pm$0.05 & 80.24$\pm$0.09 & 98.23$\pm$0.10 & 90.71$\pm$0.16 & \textbf{87.77} \\ \midrule
\multirow{3}{*}{16shot} & TaskRes & Standard & 80.46$\pm$0.01 & 72.03$\pm$0.46 & 93.72$\pm$0.13 & 79.92$\pm$0.13 & 98.00$\pm$0.08 & 91.47$\pm$0.05 & 85.93 \\
 & TaskRes & Base+CuPL & 81.78$\pm$0.02 & 81.28$\pm$0.82 & 99.22$\pm$0.12 & 79.92$\pm$0.17 & 98.47$\pm$0.08 & 91.19$\pm$0.11 & \textbf{88.65} \\
 & SToRI (Ours) & Base+CuPL & 81.40$\pm$0.02 & 79.89$\pm$0.70 & 98.58$\pm$0.06 & 81.43$\pm$0.16 & 98.47$\pm$0.12 & 91.25$\pm$0.04 & \underline{88.50} \\
 \bottomrule
\end{tabular}
\end{adjustbox}
\endgroup
\caption{Accuracy (\%) on few-shot classification with MetaCLIP ViT-L/14. The results include mean values with Standard deviation across three runs. The results of TaskRes are reproduced. The best performance is indicated in bold, while the second-best performance is underlined.}
\label{tab:zeroshot_perf_metaclip}
\end{table*}

%% file: tabs/tab_celeba_textprompt.tex
\begin{table*}
\centering
\begin{adjustbox}{max width = 1.0\textwidth}
\begin{tabular}{ll}
\toprule
Selected Attributes & Text prompts \\ 
\midrule
Female/Male, Smiling, Bangs & a photo of a smiling [woman/man] with bangs \\
Female/Male, Smiling, Blond Hair & a photo of a smiling [woman/man] with blond hair \\
Female/Male, Smiling, Gray Hair & a photo of a smiling [woman/man] with gray hair \\
Female/Male, Smiling, Wearing Hat & a photo of a smiling [woman/man] wearing hat \\
Female/Male, Smiling, Eyeglasses & a photo of a smiling [woman/man] wearing eyeglasses \\
Female/Male, Bangs, Wearing Hat & a photo of a [woman/man] with bangs, wearing hat \\
Female/Male, Bangs, Eyeglasses & a photo of a [woman/man] with bangs, wearing eyeglasses \\
Female/Male, Blond Hair, Eyeglasses & a photo of a [woman/man] with blond hair, wearing eyeglasses \\
Female/Male, Gray Hair, Eyeglasses & a photo of a [woman/man] with gray hair, wearing eyeglasses \\
Female/Male, Wearing Hat, Eyeglasses & a photo of a [woman/man] wearing hat and eyeglasses \\
 \bottomrule
\end{tabular}
\end{adjustbox}
\caption{All combinations of attributes and corresponding text prompts on the CelebA dataset.}
\label{tab:text_prompt_celeba}
\end{table*}

\begin{table*}
\centering
\begin{adjustbox}{max width = 1.0\textwidth}
\begin{tabular}{ll}
\toprule
Attributes & Texts \\ 
\midrule
has\_bill\_shape::hooked\_seabird & hooked seabird bill\\
has\_shape::duck-like & duck-like shape\\
has\_crown\_color::blue & blue crown \\
has\_forehead\_color::blue & blue forehead \\
has\_wing\_color::yellow & yellow wing \\
upperparts\_color::yellow & yellow upperparts \\
has\_underparts\_color::yellow & yellow underparts \\ 
has\_back\_color::yellow & yellow back \\
has\_breast\_color::yellow & yellow breast \\
has\_throat\_color::yellow & yellow throat \\
has\_forehead\_color::yellow & yellow forehead \\
has\_nape\_color::yellow & yellow nape \\
has\_belly\_color::yellow & yellow belly \\
has\_primary\_color::yellow & yellow color  \\
has\_crown\_color::yellow & yellow crown \\

 \bottomrule
\end{tabular}
\end{adjustbox}
\caption{Candidates of attributes and corresponding texts on the CUB dataset.}
\label{tab:text_prompt_cub}
\end{table*}

%% file: tabs/fig_ablation_layer.tex
\begin{figure*}[t]
\begin{center}
\centerline{\includegraphics[width=\linewidth]{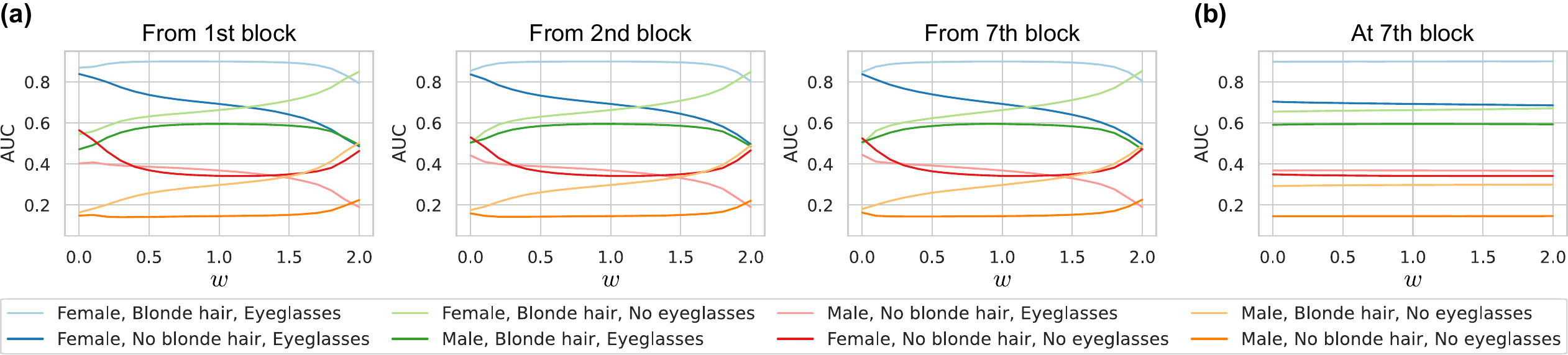}}
\caption{The change of AUC scores for preference retrieval with weight control when diversifying blocks that semantic token reweighting is applied. (a) The results when reweighting is applied within the subsequent blocks as well. (b) The result when reweighting is applied within a single block.}
\label{fig:ablation_layer_retrieval}
\end{center}
\end{figure*}

%% file: tabs/fig_ablation_layer_fewshot.tex
\begin{figure*}[t]
\begin{center}
\centerline{\includegraphics[width=0.9\linewidth]{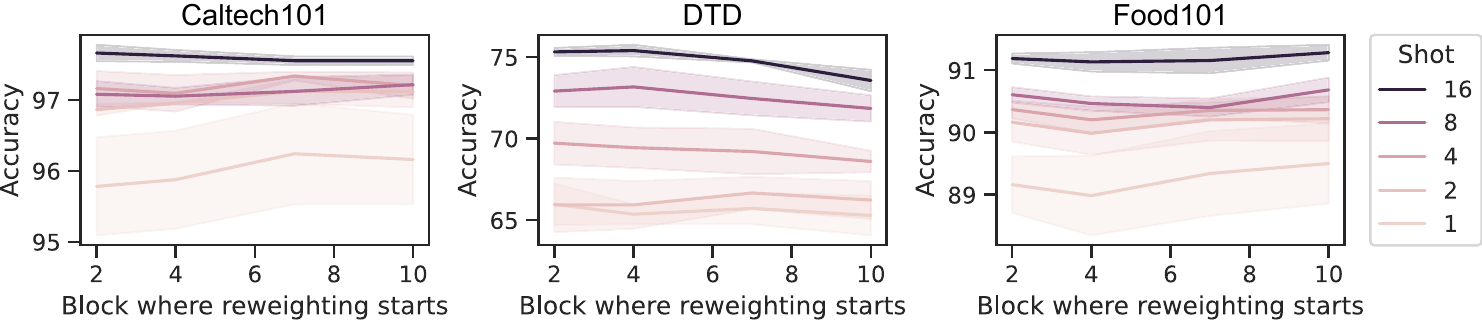}}
\caption{The change of accuracy for few-shot classification when diversifying blocks that semantic token reweighting is applied. The experiments are run three times, with the mean shown by a line and the standard deviation indicated by shading.}
\label{fig:ablation_layer_fewshot}
\end{center}
\end{figure*}